\pdfoutput=1
\documentclass[10pt]{article}

\usepackage[T1]{fontenc}
\usepackage[utf8]{inputenc}
\usepackage{lmodern}

\usepackage{amsmath}
\usepackage{amssymb}
\usepackage{amsthm}

\usepackage{booktabs}
\usepackage{graphicx}
\graphicspath{{figures/}}

\usepackage[margin=1in,marginparwidth=0.9in]{geometry}
\usepackage{microtype}
\usepackage{tabularx}
\usepackage{tikz}
\usetikzlibrary{arrows.meta, positioning}

\usepackage[numbers]{natbib}
\usepackage{hyperref}
\hypersetup{colorlinks=true, linkcolor=blue, citecolor=blue, urlcolor=blue}

\newtheorem{lemma}{Lemma}

\theoremstyle{definition}
\newtheorem{result}{Result}

\newcommand{\evtag}[1]{\textit{[#1]}}

\newcommand{\figorbox}[2]{\includegraphics[width=#2]{#1}}

\title{Detecting a Route Flip Is Easier Than Knowing Whether to Fix It:\\
Causal Route-Mediated Damage in Quantized Mixture-of-Experts}

\author{Parvel Gu\thanks{Corresponding author and submitter: \texttt{parvel.gu@gmail.com}} \\
  Independent Researcher \\
  \texttt{parvel.gu@gmail.com}}

\date{\today}

\begin{document}
\maketitle

\begin{abstract}
Top-$k$ Mixture-of-Experts (MoE) routing is discontinuous, so a deployment-motivated
numerical disturbance---simulated 4-bit KV-cache quantization read by a protected BF16
gate---pushes tokens across decision boundaries and flips which experts fire. This paper
proposes no new mitigation; it supplies a causal apparatus, several empirical findings,
and a detection-limit result for that disturbance. A four-run apparatus prices the
route-mediated fraction (RMF) of quantization damage, a token-level attribution
decomposes it by mechanism, and pre-registered probes carry the findings across three
architectures. On OLMoE-1B-7B at 4-bit KV (pilot), about a third of the damage is
routing-mediated under quantized compute: the route-mediated fraction is
$\mathrm{RMF}\approx0.31$, confirmed as the process-replicated central value (five
independent process invocations, mean $0.313\pm0.020$), with the discovery anchor
$0.31$ $[0.20,0.41]$ and a pre-registered re-execution at $0.231$. The deployable router
margin detects that a flip occurred (AUC $0.772$) but cannot tell a harmful flip from a
helpful one (at chance): among the tested local, inference-observable router statistics we
find no predictor of a flip's loss sign above chance---an empirical benefit-detection
barrier bounding selective repair restricted to this feature family. The signed-flip tax
and the sign-inseparability carry cross-model; the clean-reference remedy's payout is
architecture-modulated; a controlled same-checkpoint flag-swap re-scopes the gate's
normalization convention to a \emph{damage-magnitude} moderator, not a route-recoverability
mechanism. A real int4 KV kernel
yields a route-mediated fraction compatible with the frozen fake-quant dose curve but is
underpowered (95\% CI $[-0.111,0.394]$ includes zero): it rules out gross disagreement at
int4, but is not an independent replication. Headline hypotheses,
thresholds, and evaluations were pre-registered before measurement, with misses reported.
A pre-registered held-out read replicates the partition and the near-cancelling tax out of
sample; the strict impossibility exclusion narrowly misses.
\end{abstract}

\section{Introduction and three claims}
\label{sec:intro}
Hard top-$k$ expert routing scaled through \citet{gshard}, \citet{switch},
\citet{expertchoice} and stability-regularized \citep{stmoe} variants into deployed
sparse-MoE LLMs \citep{mixtral}; routers encode their experts' geometry
\citep{ahrac2026}. Selection by $\arg\text{top-}k$ is discontinuous, so a small numerical
perturbation of the router's input can change \emph{which} experts fire. We study one
deployment-origin disturbance---4-bit KV-cache quantization \citep{kvquant, kivi} under a
\emph{protected} BF16 gate (router weights and arithmetic in BF16, reading a quantized
upstream hidden state)---and ask what damage flows through the routing decision, and
whether that damage can be detected or repaired from signals available at inference.

This paper contributes a measurement apparatus and three findings, not a method:

\begin{enumerate}\itemsep2pt
\item \textbf{A priced routing channel.} A four-run causal apparatus decomposes
quantization damage into a compute path and a routing path. On OLMoE at 4-bit KV, the
route-mediated fraction is $\mathrm{RMF}\approx0.31$---confirmed as the process-replicated
central value (five-process mean $0.313\pm0.020$; discovery anchor $0.31$ $[0.20,0.41]$,
pre-registered re-execution $0.231$); the compute$\times$route
interaction is \emph{measured} (not assumed additive) and excludes zero. Nearly all
($99.8\%$) of the \emph{net signed} route-mediated contribution is associated with a
route-set change at the scoring or an upstream layer (majority nonlocal)---a signed ratio
of near-cancelling components. The partition is a \emph{function} of
architecture, dose, and damage domain---not a constant. \evtag{empirical, pilot}
\item \textbf{Detecting a flip is not knowing whether to fix it.} The deployable router
margin scores flip \emph{occurrence} at AUC $0.772$ but flip \emph{harm} at chance (AUC
$0.490$). Across the tested local feature families, no predictor of a flip's loss sign
beats a rate-matched random selector---an empirical benefit-detection barrier that bounds
selective repair \emph{restricted to this feature family and protocol}. \evtag{empirical,
pilot}
\item \textbf{Reference-fidelity is architecture-modulated.} Pinning a clean route
recovers a bounded slice of the routed ceiling ($+0.23$ to $0.46$ on two architectures); a
controlled same-checkpoint flag-swap re-scopes the gate's normalization convention to a
\emph{damage-magnitude} moderator, not a route-recoverability mechanism (a fourth-model
axis is the remaining $n=3$ confound).
\evtag{empirical, pilot}
\end{enumerate}

\textbf{Claim perimeter.} We do \emph{not} show that MoE routing is a control system, that
quantization is uniquely harmful to MoE, or that route repair is impossible---only that
\emph{per-token sign-selective} repair of quantization-induced flips is bounded near
random under a protected-gate KV disturbance, at pilot scale, on the architectures
measured. We report \emph{findings}, not laws, and an \emph{empirical barrier}, not a
theorem: the negative result is scoped to the tested inference-observable statistics---which
now include a cross-layer router vector over all $16$ layers (Table~\ref{tab:probes}), also
at chance---and does not rule out predictors using richer hidden-state or trained-decoder
information. Every claim carries an evidence-grade tag \evtag{empirical/pilot/bound}; the reserved held-out split is read
\emph{once}, on three pre-registered endpoints (\S\ref{sec:extval}), with the one miss
reported.

\section{Related work and experimental setting}
\label{sec:related}
\textbf{Discontinuity geometry.} \citet{tranhuu2026} classify routing discontinuities and
propose input smoothing evaluated on clean benchmarks with no quantization study; we test
that mechanism under the KV disturbance. \textbf{Causal audit of routing.}
\citet{engmann2026} audit whether population routing summaries predict the causal
importance of expert ablations across OLMoE, Qwen1.5-MoE and DeepSeek and find
observational metrics insufficient; we instead intervene on the route \emph{mediator}
under paired clean/quantized compute and ask whether inference-observable signals predict
the \emph{sign} of route-mediated token loss. \textbf{Router-aware MoE PTQ.}
\citet{eaquant2025}, \citet{vsraq2026}, \citet{gemq}, \citet{routercalib2026} align
router logits under static weight/activation quantization; none reaches the KV disturbance
or prices the damage causally. \textbf{Soft/continuous routing.} \citet{softmoe2026}
trains relaxations with no quantization claim; routing-free MoE \citep{routingfree2026}
removes the router entirely. \textbf{KV compression / allocation.} GEAR-class
reconstruction \citep{gear} targets the attention output, not the routing argmax;
layer-wise precision pairs loss-sensitivity to bits \citep{kvtuner2025, kvmix2026,
mixkvq2026}; rotation-based W4A4 \citep{quarot, duquant} is orthogonal to the KV axis. On
OLMoE---the common evaluation model---our causal decomposition refines the router-aware
PTQ line's correlational phenomenology: they observe that quantization perturbs routing;
we price it (the route-mediated channel carries $\approx 31\%$ of the damage).

\textbf{Setting.} Model: OLMoE-1B-7B (\texttt{allenai/OLMoE-1B-7B-0924}, pinned revision),
16 layers, 64 experts, top-$k=8$, \texttt{norm\_topk\_prob = False}, protected BF16 gate
\citep{olmoe2024}. Corpus: a calibration split with a 6-bucket manifest (WikiText-103
\citep{wikitext}, C4 \citep{c4}); the test split is unread except for one pre-registered confirmatory read (\S\ref{sec:extval}). Decode: sequential
teacher-forced, 64-token prefill then decode, per-token NLL. The pilot anchor is $N=96$
sequences ($18{,}432$ decode tokens); where measured, an expanded nested-calibration
sample of $N=384$ ($4{\times}N$) is carried alongside. Cross-model probes use
Qwen1.5-MoE-A2.7B \citep{qwenmoe2024}, Qwen3-30B-A3B \citep{qwen3}, and DeepSeek-MoE-16B
\citep{deepseekmoe2024}. The primary disturbance is a \emph{simulated} 4-bit symmetric
fake-quantization of the KV cache at full strength; an INT8-activation condition is a
near-null (excess NLL $\approx 0$), so the damaging channel is the KV cache read
upstream of the router, through attention.

\section{Four-run causal apparatus and estimands}
\label{sec:apparatus}
\textbf{Four runs} (single process per contrast):
$Y_{CC}$ (clean compute, free route; baseline);
$Y_{QP}$ (quantized compute, clean route \textbf{pinned});
$Y_{QF}$ (quantized compute, free route);
$Y_{CT}$ (clean compute, quantized route \textbf{transplanted}).
Pinning imposes the clean expert \emph{set} \emph{and} its clean weights verbatim (no
renormalization, matching \texttt{norm\_topk\_prob = False}); this is an
\emph{interchange intervention} on the routing mediator in the activation-patching lineage
\citep{rome2022, vig2020, geiger2021, pathpatch2023, ioi2022, attrpatch2023}, not an
ablation. \textbf{Estimand semantics.} RMF is an \emph{interventional controlled-route
contribution under quantized compute, not a natural indirect effect}: the pinned condition
is intentionally cross-context (a clean route set and its clean weights read against a
quantized hidden state) and prices the route channel under a specified mediator
intervention; the reported conclusions depend on this intervention semantics.

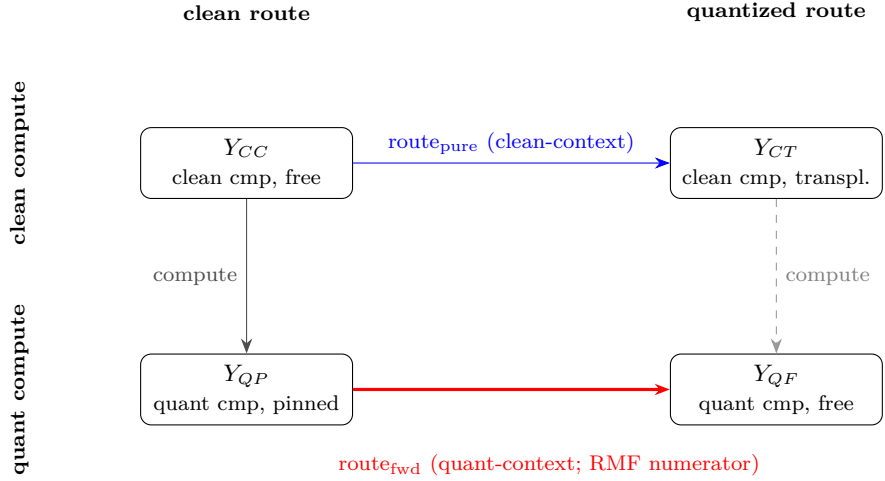
\begin{figure}[t]\centering
\begin{tikzpicture}[
  >={Stealth[length=2mm]},
  run/.style={draw, rounded corners, align=center, inner sep=4pt, font=\small, minimum width=28mm},
  lbl/.style={font=\footnotesize}, hd/.style={font=\footnotesize\bfseries}]
  \node[hd] (hc) at (0,2) {clean route};
  \node[hd] (hq) at (7,2) {quantized route};
  \node[run] (cc) at (0,0)   {$Y_{CC}$\\\footnotesize clean cmp, free};
  \node[run] (ct) at (7,0)   {$Y_{CT}$\\\footnotesize clean cmp, transpl.};
  \node[run] (qp) at (0,-3)  {$Y_{QP}$\\\footnotesize quant cmp, pinned};
  \node[run] (qf) at (7,-3)  {$Y_{QF}$\\\footnotesize quant cmp, free};
  \node[hd, rotate=90] at (-3,0)  {clean compute};
  \node[hd, rotate=90] at (-3,-3) {quant compute};
  \draw[->, blue] (cc) -- node[lbl, above, blue]{route$_\text{pure}$ (clean-context)} (ct);
  \draw[->, red, very thick] (qp) -- (qf);
  \node[lbl, red] at (4,-4) {route$_\text{fwd}$ (quant-context; RMF numerator)};
  \draw[->, black!70] (cc) -- node[lbl, left, black!70]{compute} (qp);
  \draw[->, black!40, dashed] (ct) -- node[lbl, right, black!50]{compute} (qf);
\end{tikzpicture}
\caption{The four-run apparatus as a 2$\times$2 causal graph (rows: clean vs.\ quantized
\emph{compute}; columns: clean vs.\ quantized \emph{route}). The deployment-relevant
quant-context route contrast route$_\text{fwd}=L(Y_{QF})-L(Y_{QP})$ (red, the RMF
numerator) and the clean-context route contrast route$_\text{pure}=L(Y_{CT})-L(Y_{CC})$
(blue) differ by the measured compute$\times$route interaction
(Table~\ref{tab:interaction}); the vertical edge is the compute channel
$L(Y_{QP})-L(Y_{CC})$. $\mathrm{RMF}=\sum\text{excess\_route}/\sum\text{excess\_total}$
with excess\_route $=$ route$_\text{fwd}$.}
\label{fig:fourrun}
\end{figure}

\textbf{RMF estimator.}
$\text{excess\_total}=L(Y_{QF})-L(Y_{CC})$;
$\text{excess\_compute}=L(Y_{QP})-L(Y_{CC})$;
$\text{excess\_route}=\text{excess\_total}-\text{excess\_compute}$; and
$\mathrm{RMF}=\sum\text{excess\_route}/\sum\text{excess\_total}$---a \emph{ratio of sums},
never a median of per-sequence ratios (unbounded variance near zero denominators). CIs are
by \textbf{sequence-cluster bootstrap} \citep{efron1993} (resample sequences, not tokens);
combined-logistic AUCs are \emph{sequence-held-out} train/val splits, not train-fits. An
RMF is interpretable only when $\text{excess\_total}>0$ and its CI excludes zero. The two
route paths are \emph{not} additively separable at this dose: the interaction is measured
($Y_{CT}$ supplies the clean-context path), not assumed away.

\textbf{Token-level attribution.} For each decode token we capture the top-$k$ route set at
every layer under clean and quantized compute, set
$\mathrm{harm}(t)=L(Y_{QF})(t)-L(Y_{QP})(t)$, and classify: \textbf{jump} if the scoring
layer's set changed; else \textbf{nonlocal} if any upstream layer's set changed; else
\textbf{pure flux}. The three masks partition all tokens, so $\sum\mathrm{harm}$
partitions exactly into jump $+$ nonlocal $+$ pure-flux. \textbf{Estimand caveat.} Under
the blanket intervention the token-level harm label is an \emph{attribution label}---the
sign of that token's counterfactual $\Delta$NLL, quantized vs.\ clean route---\emph{not} a
composable individual treatment effect. The utility of a selective strategy is therefore
evaluated by \emph{running} that strategy (\S\ref{sec:detection}), never by summing labels.

\section{RMF, interaction and mechanism partition}
\label{sec:rmf}
\begin{table}[t]\centering\small
\caption{Two-path decomposition of the route effect (4-bit KV, OLMoE,
$N{=}96$). The quant-context path (route$_\text{fwd}$, the RMF numerator) exceeds the
clean-context path; their difference is the measured compute$\times$route interaction $I$
(CI excludes 0)---itself a deployment fact, since a route-only remedy cannot be budgeted as
if compute damage were held fixed. This re-run's route fraction
(route$_\text{fwd}/$total $=0.0175/0.0757=0.231$) lies inside the headline RMF CI
$[0.20,0.41]$; the gap to the $0.31$ anchor is run-to-run variation and winner's curse.}
\label{tab:interaction}
\begin{tabular}{@{}l l@{}}\toprule
path & nats (95\% CI) \\ \midrule
route$_\text{fwd}$ $= L(Y_{QF})-L(Y_{QP})$ (quant-context; RMF numerator) & $+0.0175$ $[0.0113, 0.0234]$ \\
route$_\text{pure}$ $= L(Y_{CT})-L(Y_{CC})$ (clean-context) & $+0.0092$ $[0.0048, 0.0134]$ \\
\textbf{interaction} $I =$ route$_\text{fwd}-$route$_\text{pure}$ & $\mathbf{+0.0083}$ $\mathbf{[+0.0039, +0.0124]}$ \\
\midrule
INT8 near-null control (interaction scales with damage) & $-0.00006$ \\
\bottomrule
\end{tabular}
\end{table}

\begin{result}[Route-mediated fraction and mechanism partition]
On OLMoE at 4-bit KV, $\mathrm{RMF}=0.31$, 95\% CI $[0.20,0.41]$ (excludes zero). This is
the \emph{process-replicated} central value: over five independent process invocations the
four-run RMF mean is $0.313\pm0.020$ (range $0.288$--$0.339$; \S\ref{sec:extval},
Table~\ref{tab:xproc}), the discovery anchor is $0.31$, a pre-registered re-execution gives
$0.231$, and fixed-process re-estimates range $\approx0.23$--$0.33$. The damage
partitions \textbf{jump $45\%$ $\cdot$ nonlocal $55\%$ $\cdot$ pure-flux $\approx0.2\%$};
nearly all ($99.8\%$) of the \emph{net signed} route-mediated contribution is associated
with a route-set change at the scoring or an upstream layer (majority nonlocal). This share
is a ratio of \emph{signed}
per-token contributions; because the oppositely-signed flip components near-cancel
(harmful $+0.043$ / beneficial $-0.047$ nats, net $\approx 0$; \S\ref{sec:detection}), it
is reported as the signed set-change share, not an absolute-value partition. A
single-scoring-layer indicator localizes only $45\%$; the attribution algorithm recovers
the rest as upstream jumps. \evtag{empirical, pilot}
\end{result}

\textbf{The partition is a function, not a number.} It is a mapped function of architecture
$\times$ dose (an aggressive-dose sweep, re-estimated at $N{=}384$; App.~\ref{app:extras})
$\times$ damage domain: in the KV domain $\mathrm{RMF}=0.31$; in the weight-PTQ domain
\citep{gptq, awq, smoothquant} a first-party \emph{conservative lower bound} is
$0.115$--$0.123$, stable across a $\approx 8\times$ dose range (expert-FFN RTN at 4-bit and
3-bit). The weight-domain lower bound is a different damage domain \emph{and} dose from the
KV pilot anchor $0.31$, and is not a re-measurement of it. \evtag{empirical, pilot}

\section{Detection versus selective utility}
\label{sec:detection}
Let $m$ be the quantized router margin (weakest-selected minus strongest-unselected
logit), the strongest deployable single-layer statistic at inference. We distinguish four
objects: $F_\text{score}$, a route-set change at the \emph{scoring} layer (where the output
NLL is read); $F_\text{any}$, a route-set change at \emph{any} layer (scoring or upstream);
$H$, the token attribution sign under the blanket route intervention (whether pinning the
clean route lowers this token's loss); and $B_\pi$, the realized benefit of \emph{executing}
a selective policy $\pi$. A pin helps token $t$ iff $F_\text{any}$ holds and $H$ is
favorable. The scoring-layer margin predicts $F_\text{score}$, whereas selective benefit
$B_\pi$ depends on a trajectory-level functional involving $F_\text{any}$ and downstream
state---so a high flip-occurrence AUC ($0.772$, an $F_\text{score}$ read-out) and a large
$P(\text{harm}{>}0\mid\text{no scoring-layer flip})=0.55$ are consistent, not conflicting.

\begin{result}[Benefit-detection barrier (empirical)]
From two measured premises: (P1) the single-layer margin scores flip \emph{occurrence} at
$\text{margin}\to\text{flip}$ AUC $=0.772$; (P2) given a flip, the margin does not resolve
harm---$\text{margin}\to(\text{harmful}\mid\text{flip})$ AUC $=0.490$,
$P(\text{harmful}\mid\text{flip})=0.572$ (a near coin flip). Because the margin carries the
first conjunct but not the second, a benefit predictor built from the tested local
statistics scores at the barrier: benefit-vs-all AUC $=0.499$. \evtag{empirical}
\end{result}

\textbf{Population semantics.} The three AUCs are on distinct populations: flip-vs-all
tokens ($0.772$), harmful-given-flip \emph{within} flipped tokens ($0.490$), and
benefit-vs-all tokens ($0.499$). The last is a mixture over two negative groups---non-flip
tokens (weight $0.576$, separable at AUC $0.615$) and flipped-but-harmless tokens (weight
$0.424$, AUC $0.342$)---reconstructing $0.576\times0.615+0.424\times0.342=0.499$. Benefit
is not nested in the single-layer flip indicator ($P(\text{harm}{>}0\mid\text{no
flip})=0.55$; non-flip contribution $+0.009$ $[+0.005,+0.013]$ nats) precisely because the
damage is majority nonlocal: a single-layer flip cannot carry a multi-layer harm. Richer
probes agree: nonlinear and temporal predictors of harm-given-flip (MLP, gradient-boosted
trees, and a budget-matched temporal predictor) all sit at chance, sequence-held-out. The
combined-logistic sign-AUC is $0.520$ (OLMoE), within the pre-declared
$|\mathrm{AUC}-0.5|\le 0.05$ falsification band.

\begin{figure}[t]\centering
  \figorbox{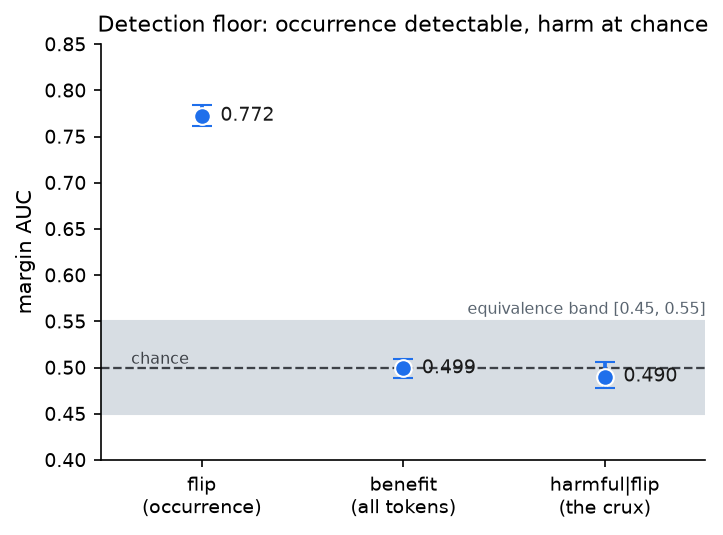}{0.6\linewidth}
  \caption{The benefit floor. Deployable router margin:
  flip-detection AUC $0.772$ vs harmful-flip localization AUC $0.490$ (benefit-prediction
  AUC $0.499$). The margin carries occurrence but not harm-direction.}
  \label{fig:margin}
\end{figure}

\begin{table}[t]\centering\small
\caption{Oracle-vs-random battery (clean-route-pin instrument,
both architectures, $N{=}384$; fractions of the route-mediated component). The
oracle-omniscient advantage (oracle $-$ random) is null on OLMoE and positive on DeepSeek;
the deployable margin gate recovers below random on both. This is a \emph{measured
cross-model contrast}, not a missing control.}
\label{tab:orc-battery}
\begin{tabular}{@{}l c c@{}}
\toprule
Gate (fraction of harmful) & OLMoE & DeepSeek \\
\midrule
oracle & $8.9\%\,[6.4,11.4]$ & $8.4\%\,[6.4,10.6]$ \\
rate-matched random & $8.4\%\,[5.7,11.1]$ & $6.1\%\,[4.1,8.1]$ \\
deployable margin & $3.6\%\,[1.4,5.9]$ & $4.3\%\,[2.8,5.7]$ \\
\textbf{oracle $-$ random (paired)} & $\mathbf{+0.4\%\,[-1.8,+2.8]}$ (incl 0) & $\mathbf{+2.3\%\,[+0.9,+3.7]}$ (excl 0) \\
margin $-$ random (paired) & below random & $-1.8\%\,[-3.5,-0.1]$ (excl 0) \\
\bottomrule
\end{tabular}
\end{table}

\textbf{Selective utility, run not summed.} We evaluate selective policies by executing
them. An \emph{oracle} gate (given the true harm sign) and a rate-matched \emph{random}
gate bound the achievable recovery; the deployable \emph{margin} gate is the realistic
policy (Table~\ref{tab:orc-battery}). On OLMoE the oracle buys nothing over random
($+0.4\%$ $[-1.8,+2.8]$, CI includes 0); on DeepSeek an oracle label buys a small but
significant $+2.3\%$ $[+0.9,+3.7]$ over random. The \emph{deployable} margin gate recovers
\emph{below} random on both architectures ($-1.8\%$ $[-3.5,-0.1]$ on DeepSeek, precision
$\approx$ chance). So the ``oracle beats nothing over random'' form is OLMoE-specific and
did not generalize, while the practical floor---no deployable recovery from local
features---holds on both. \evtag{empirical, pilot}

\textbf{Design implication (scoped).} An inference-time route correction driven by the
\emph{tested} deployment-observable features---including online token-wise router
calibration built from those statistics---inherits the barrier for harmful-flip
localization. This bounds selective-repair schemes \emph{restricted to this feature family
and evaluation protocol}---now including a cross-layer router vector over all $16$ layers,
also at chance (Table~\ref{tab:probes})---and does not rule out predictors using richer
hidden-state or trained-decoder information. To refute the floor, a method must beat random at
locating harmful-to-pin tokens using \textbf{only inference-observable features}, with our
labels, budget-matched, on a paired CI vs random that excludes zero. \evtag{empirical}

\section{Cross-model replication and a controlled normalization flag-swap}
\label{sec:crossmodel}
\textbf{The tax and the sign-inseparability carry.} On DeepSeek-MoE-16B ($N{=}384$,
dose-matched), route-mediated damage decomposes into harmful ($+0.2036$ $[+0.199,+0.208]$)
and beneficial ($-0.1819$ $[-0.186,-0.178]$) flip components that near-cancel
(\textbf{cancellation $90.2\%$}), and the oracle harm sign is not predictable from local
router statistics (combined sign-AUC $0.507$, within the $|\mathrm{AUC}-0.5|\le0.05$
band). The beneficial-flip tax and sign-inseparability therefore hold on a second
architecture.

\begin{table}[t]\centering\small
\caption{Preliminary cross-architecture association for
prefix-reference recovery (all rows $N{=}384$, nested). The Qwen3 row is the
\emph{prefix-only} deployable reference; a controlled full-decode contrast recovers on
Qwen3 (below), localizing the null to reference semantics. The normalization-convention
mechanism was registered \emph{before} the DeepSeek measurement; the DeepSeek row is a
genuine prediction-before-measurement, the OLMoE and Qwen3 rows predate it.}
\label{tab:threemodel}
\begin{tabular}{@{}l l p{4.6cm}@{}}
\toprule
model & \texttt{norm\_topk\_prob} & clean-reference recovery (fraction of routed ceiling) \\
\midrule
OLMoE            & \texttt{False} & $+0.231$ $[0.095, 0.363]$ \\
DeepSeek-MoE-16B & \texttt{False} & $0.456$ $[0.246, 0.597]$ \\
Qwen3-30B        & \texttt{True}  & $0.021$ $[-0.084, 0.116]$ (null) \\
\bottomrule
\end{tabular}
\end{table}

\textbf{Reference-fidelity, re-estimated.} Pinning a clean prefix route recovers a bounded
directional slice of the routed ceiling: discovered $+0.348$ $[0.076,0.622]$ at $N{=}96$,
\emph{re-estimated} $+0.231$ $[0.095,0.363]$ on the expanded nested sample ($4{\times}N$;
the interval tightened $\times0.49\approx 1/\sqrt{4}$, CI excludes zero). We report this
as measured with \emph{no dominance claim}. The payout is architecture-modulated
(Table~\ref{tab:threemodel}): the two \texttt{False}-class magnitudes (OLMoE $+0.231$,
DeepSeek $0.456$) have overlapping intervals ($[0.246, 0.363]$), so we claim no rank
ordering; the \texttt{True}-class architecture is null under the deployable prefix-only
reference.

\textbf{Candidate moderator, controlled contrast, and firewalls.} A same-checkpoint,
adapter-identity-verified contrast (max$|\Delta\text{NLL}|=0$) isolates the Qwen3 null to
the \emph{deployable prefix-only} reference: under full-decode clean-route pinning, Qwen3
recovers $+0.408$ $[+0.363, +0.574]$ of \emph{total} quant damage versus the prefix-only
null $+0.038$, with an OLMoE control at $+0.271$. \textbf{Denominator firewall:} the
clean-reference recovery $+0.231$ is a fraction of the \emph{route-mediated} ceiling (quant
vs.\ all-clean-routes-pinned), whereas the full-decode/prefix-only recoveries are fractions
of \emph{total} quant damage (quant vs.\ clean)---the same intervention family under
different denominators, not directly comparable. The convention thus appears to moderate
the deployable-reference case specifically---but a controlled test now shows this is
\emph{not} a route-recoverability mechanism. \textbf{Controlled flag-swap (the decisive
single-variable test).} On a fixed checkpoint we force the opposite \texttt{norm\_topk\_prob}
and re-measure the clean-route pin recovery, with the adapter-identity gate PASSING
(max$|\Delta\text{NLL}|=0$) on all four native/forced $\times$ \{OLMoE, Qwen3\} configs---a
genuinely single-variable swap. Clean-route full-decode recovery \emph{persists under both
conventions on both checkpoints} (OLMoE $+0.288$/$+0.483$, Qwen3 $+0.408$/$+0.596$; all four
CIs exclude zero; the toggled configs are off-native-distribution, so we read robustness, not
the magnitude delta). The one clean single-variable effect of the convention is on the
\emph{damage magnitude} (routed ceiling $\times7$--$10$, \texttt{False}${=}$more damage), not
recoverability.

\emph{Synthesis (P7).} The pre-registered discriminator pattern stands as an association
under the prefix protocol. Two controlled single-variable tests refine its interpretation:
the convention bit does not gate recoverability (recovery persists under both conventions,
both checkpoints), and the original null is protocol-scoped (full-decode transplant recovers
on the \texttt{=True} model). The convention does causally moderate damage magnitude
($\times7$--$10$, single-variable). What modulates cross-architecture recovery magnitude
remains open ($n{=}3$). A second-model replication on Qwen1.5-MoE
gives $\mathrm{RMF}=0.290$ $[0.200, 0.372]$, with its always-on shared-expert channel
($10.6\%$ of FFN-output norm) shifting damage into a class-A exposure buffer.
\evtag{empirical, pilot, 3 models}

\section{Real-kernel and held-out validation}
\label{sec:extval}
\textbf{Real int4 kernel.} To test whether the route-mediated fraction is a simulation
artifact, a real low-bit KV kernel (a pack-to-int4 / dequant-on-read HQQ backend) replaces
the fake-quant disturbance. In the same session the positive control reproduces the
canonical partition (fake-quant RMF $0.339$ $[0.247,0.427]$, covering the $0.31$ anchor),
and the \textbf{real int4 RMF is $0.194$ $[-0.111,0.394]$}. The point estimate is
\emph{compatible} with the band predicted from the frozen fake-quant dose curve
($[0.087,0.350]$ at the measured harm), but the experiment is \emph{underpowered}: the
95\% CI includes zero, so it rules out gross disagreement at int4 without constituting an
independent replication of a nonzero route-mediated fraction. \textbf{Dose mismatch (stated
honestly):} the real int4 kernel inflicts total damage $0.0082$ nats versus the
fake-quant $1\times$ operating point's $0.0757$ nats---the same nominal bit-width at a very
different \emph{damage dose}, so this is not the fake-quant ``headline dose'' point.
\textbf{Equal-prominence divergence:} the real-kernel RMF \emph{declines} with dose
($0.194\to0.129\to0.090$ for int4$\to$int3$\to$int2) while the fake-quant curve
\emph{rises}, so at the aggressive int2 dose the real share falls \emph{below} its
predicted band---the optimized kernel loads more damage onto the direct channel at
aggressive bits---itself a measurement, not a refutation. \evtag{empirical, pilot}

\textbf{Process replication.} To answer whether the RMF numerator---a small per-token
quantity---survives a change of process (a different reduction order / kernel schedule), we
re-ran the three core paired contrasts once per fresh Python process across five
independent invocations (OLMoE, kv4/s1.0, $N{=}96$; Table~\ref{tab:xproc}). Within a
process the paired execution is bitwise-identical (control max$|\Delta|=0$); across
processes the four-run RMF has mean $0.313$ with between-process SD
$\sigma_{\mathrm{proc}}=0.020$ (range $0.288$--$0.339$), i.e.\ $\mathrm{RMF}=0.31\pm0.02$,
and the per-token route-numerator between-process SD is $\sigma_{\mathrm{proc}}\le0.0018$
nats. All variance is between-process ($\text{ICC}_{\text{process}}=1.0$; within-process
$0$). \textbf{The route numerator ($\approx0.016$ nats) sits an order of magnitude above
the measured cross-process floor ($\le0.0018$ nats)}, so RMF is \emph{not} a
single-process reduction-order artifact; the $0.313$ mean confirms $0.31$ as the
process-central value rather than a winner's-curse high. \evtag{empirical, pilot}

\begin{table}[t]\centering\small
\caption{Process replication: five fresh process invocations of the core contrasts (OLMoE,
kv4/s1.0, $N{=}96$). $\sigma_{\mathrm{proc}}$ is the between-process SD; the within-process
control is bitwise ($\max|\Delta|=0$), so $\text{ICC}_{\text{process}}=1.0$. The per-token
route-numerator floor ($\le0.0018$ nats) is an order of magnitude below the route numerator
($\approx0.016$ nats): the RMF is process-stable, not a reduction-order artifact.}
\label{tab:xproc}
\begin{tabular}{@{}l c c c@{}}\toprule
endpoint (5 processes) & mean & $\sigma_{\mathrm{proc}}$ & range \\ \midrule
four-run RMF (ratio) & $0.313$ & $0.020$ & $0.288$--$0.339$ \\
route numerator (nats/tok) & $0.016$ & $0.0013$ & --- \\
within-process control (nats) & $0$ & $0$ & --- \\
\bottomrule
\end{tabular}
\end{table}

\textbf{Task-level reach.} On a task-shaped workload (ARC-Easy, $N{=}500$) the KV-quant
route tax reproduces in NLL and in its signed-flip cancellation signature (cancellation
$0.966$, harmful-dominant), but does not produce a resolvable answer-accuracy drop---the
task-level teeth are at most a $\lesssim 1$-point directional echo. We therefore read this
as an \emph{inference-mechanism and loss-decomposition} result, \emph{not} evidence of a
large downstream accuracy impact at this scale; the abstract and introduction claim no
strong deployment harm while the task-level effect is unresolved. \evtag{empirical, pilot}

\textbf{Held-out confirmation (the reserved test split, read once).} We read the reserved
\texttt{test} split \emph{once} on three pre-registered endpoints (the protocol was hashed
and committed before the read; read-only, whatever fires reported undiluted). \textbf{(1) Partition
replicates:} the four-run RMF is $0.429$ $[0.330, 0.519]$ on held-out; its CI overlaps the
discovery interval $[0.20, 0.41]$ (\textbf{MET}), and the route-mediated share is if anything
\emph{stronger} out of sample---not a winner's-curse inflation of $0.31$. \textbf{(2) Tax null
holds, with a stated caveat:} the two-sided signed-flip tax nets $+0.0043$ $[+0.0005, +0.0079]$
($\Delta|_{\text{harm}}{+}0.042$ / $\Delta|_{\text{benef}}{-}0.049$), far below $20\%$ of either
signed component (\textbf{MET} in the pre-registered equivalence sense)---the near-cancellation
confirms out of sample; undiluted, the held-out net CI \emph{excludes zero}, so ``a small,
sign-balanced residual'' survives but ``exactly zero'' does not. \textbf{(3) The strict
impossibility exclusion narrowly misses:} oracle-gated recovery is $+0.0042$ $[+0.0003,
+0.0082]$, and the pre-registered strict bound (upper $95\%$ CI $<0.008$) is \textbf{missed by
$0.0002$}. The recovery stays tiny and consistent with the impossibility \emph{direction}, but
we do \emph{not} assert a strict held-out exclusion; this miss is recorded at equal prominence
(Table~\ref{tab:ledger}). Two of three headline endpoints confirm out of sample; the strict
exclusion is softened to an unconfirmed bound. The $N{=}384$ results remain an
\emph{expanded nested calibration} re-estimate (never ``confirmed''), distinct from this read.

\section{Limitations and conclusion}
\label{sec:limits}
\textbf{Limitations.} Pilot $N$ (96 sequences, with $N{=}384$ expanded calibration where
measured); a simulated primary disturbance with a real int4 anchor whose point estimate is
compatible with the fake-quant dose curve but underpowered (95\% CI $[-0.111,0.394]$
includes zero; the aggressive-dose divergence is future work); NLL-primary metrics; a
single primary condition (4-bit KV). The RMF numerator (route$_\text{fwd}\approx0.018$
nats) is small, so the RMF is reported as a sequence-bootstrap CI \emph{conditional
on the fixed-process paired execution}; across fixed-process runs the estimated share
ranged $\approx0.23$--$0.33$. The cross-process wobble is now measured directly
(five-process replication, \S\ref{sec:extval}): $\sigma_{\mathrm{proc}}\le0.0018$ nats on
the per-token primitives---an order of magnitude below the route numerator---so the RMF is
process-stable ($\pm0.02$ on the ratio, $\text{ICC}_{\text{process}}=1.0$); only the
smaller sub-floor effects (the smoothing residual, the smallest oracle$-$random
differences) remain \emph{exploratory}. All reported contrasts are single-process and thus
paired bitwise-valid. The held-out read is done (\S\ref{sec:extval}): the partition ($0.429$) and the
near-cancelling tax replicate out of sample; the strict impossibility exclusion narrowly
misses (upper CI $0.0082$ vs the $0.008$ bar) and is recorded as a miss, not asserted.

\textbf{Conclusion.} We supply a four-run causal apparatus that prices the routing channel
of KV-quantization damage, a mechanism partition, an empirical benefit-detection barrier
for the tested local feature family, and a cross-model account of when a clean-reference
remedy pays out. We claim no new mitigation method and no state-of-the-art result: the
contribution is a measurement apparatus and a set of honestly scoped findings about what
route-mediated damage is, what can be detected of it at inference, and what reference
information its repair requires. \evtag{empirical, pilot}

\section*{Reproducibility statement}
Every measured claim traces to a committed metric JSON and a signed decision record;
model identifiers and revisions, dataset splits, decode protocol, determinism settings,
and the disturbance specification are given in \S\ref{sec:related}--\S\ref{sec:apparatus}
and App.~\ref{app:extras}. Reproducibility is within-process bitwise (identical-repeat
max$|\Delta\text{NLL}|=0$) with a cross-process wobble measured by five-process replication
($\sigma_{\mathrm{proc}}\le0.0018$ nats per-token, $\pm0.02$ on the RMF ratio,
$\text{ICC}_{\text{process}}=1.0$; \S\ref{sec:extval}) and absorbed by single-process
discipline. A
specification of the four-run causal machinery (RMF with winner's-curse anchors, the
jump/nonlocal/flux attribution, the detection-vs-benefit test, and the determinism ladder)
accompanies the paper; a full artifact release is planned for the confirmatory version.

\section*{Acknowledgments and disclosure}
AI-assisted tooling was used in research orchestration and manuscript preparation;
all experimental designs, adjudications, and claims were human-directed and
human-verified.

\bibliography{refs}

\appendix
\section{Supplementary material}
\label{app:extras}
\emph{Relegated from the body: the disturbance/topology detail, the aggressive-dose sweep,
the predictions registry, the negative-result ledger, the evidence matrix, the
supplementary instruments, and the exchange-rate arm.}

\paragraph{Disturbance specification.} The perturbation is 4-bit symmetric
fake-quantization of the KV cache at full strength. For each written key/value slice $x$,
step $\delta=\max|x|/(2^{3}{-}1)$ over the head-dim axis (per token, per head);
$x_q=\mathrm{round}(x/\delta)\cdot\delta$ (round-to-nearest, clip to $[-8,7]$), no
zero-point. K and V share the quantizer; quantization is applied on cache write at both
prefill and decode and dequantized at read. The router weights and arithmetic remain in
BF16 over the quantized upstream state (the ``protected gate''); a strength parameter
interpolates the stored tensor toward the clean tensor ($1.0$ = full quantization). The
INT8-activation condition is a near-null (excess NLL $\approx 0$). The matched $1\times$
dose is $0.0757$ nats, band $[0.0605, 0.0908]$.

\begin{lemma}[Top-$k$ selection is discontinuous]
Let the hidden state range over a \emph{connected} region $H\subseteq\mathbb{R}^{d}$ and
let $R:H\to\{\text{vertices of }\Delta(E,k)\}$ be a deterministic hard top-$k$ selector
\citep{shazeer2017}. If $R$ is non-constant it cannot be continuous everywhere on $H$:
route changes occur on score-tie surfaces where the $k$-th and $(k{+}1)$-th logits coincide
(including higher-order ties). This is a discontinuity of the \emph{selector}; the final
MoE output can still be continuous across a tie when the crossing experts carry vanishing
weight. Hence route jumps cannot be removed globally---only pushed to low-weight boundaries
(the smoothing idea) or eliminated by a gate continuous by construction.
\end{lemma}
\noindent\evtag{background property}

\paragraph{Disturbance-gain instrument.} Per-layer gain $g_l=\mathrm{median}_{\text{token}}
\|\delta x_{l+1}\|/\|\delta x_l\|$ with $\delta x=x_{\text{quant}}-x_{\text{clean}}$: raw
$\prod g_l=56.7$; normalized (relative-perturbation) $\prod\tilde g_l=1.84$, front-loaded
($\tilde g_{L0}=2.19$; interior transitions near or below unit gain; net-of-L0 contractive
$\times0.84$). Four further instruments (accumulated covariance; temporal persistence with
lag-1 $\approx0.38$, peak $0.41$; small-gain depth decay; soundness) corroborate the
event-accumulation picture (damage accrues as jump events at successive switching surfaces,
not as amplitude growth). The na\"ive ``amplifying cascade'' reading was a
relative-vs-absolute measurement artifact, corrected here.

\paragraph{Probe battery for harm-given-flip.} Beyond the single-layer margin, richer
function classes on the tested \emph{local} router statistics do not resolve harm direction
given a flip (Table~\ref{tab:probes}; harm-given-flip AUC, sequence-held-out over $96$
sequences, $n{=}2{,}553$ flipped tokens). A logistic combination, an MLP, gradient-boosted
trees, and a short route-history (temporal) predictor all include $0.5$. Critically, the
\emph{decisive} predictor the review calls for---a cross-layer router vector reading
\emph{all $16$ layers'} margins, entropies, gate-gaps, and top-$k$ gaps ($80$
features)---was captured in a pre-registered targeted recapture and also scores at the
barrier: logistic $0.512$ $[0.468,0.546]$ and gradient-boosted $0.508$ $[0.479,0.527]$,
both inside the pre-declared $|\mathrm{AUC}-0.5|\le0.05$ band with CI covering $0.5$. The
recapture reproduced the flipped-token population exactly ($n{=}2{,}553$; harmful rate
$0.584$), so this is not a power artifact. The barrier therefore holds for the \emph{whole}
tested deployable per-layer observable set, not just the single scoring-layer margin. No
cell is fabricated.

\begin{table}[t]\centering\small
\caption{Probe battery (harm-given-flip, sequence-held-out; $n{=}2{,}553$ flipped tokens,
$96$ sequences). Every tested function class---including the cross-layer router vector over
all $16$ layers---sits within the pre-declared $|\mathrm{AUC}-0.5|\le0.05$ band, with CI
covering $0.5$. No fabricated cell.}
\label{tab:probes}
\begin{tabular}{@{}l l c@{}}\toprule
predictor & input & harm-given-flip AUC (95\% CI) \\ \midrule
margin & single layer & $0.490$ \\
logistic combination & single-layer local & $0.522$ $[0.476, 0.589]$ \\
MLP (nonlinear) & single-layer local & $0.504$ $[0.451, 0.556]$ \\
gradient-boosted trees & single-layer local & $0.515$ $[0.473, 0.551]$ \\
temporal (\,$+$ route history) & local $+$ lag-1/2 & $0.522$ $[0.476, 0.589]$ \\
\textbf{cross-layer router vector} & \textbf{all-$16$-layer margin/entropy/gap ($80$-d)} & \textbf{$0.512$ $[0.468, 0.546]$} \\
\quad (gradient-boosted) & all-$16$-layer, $80$-d & $0.508$ $[0.479, 0.527]$ \\
\bottomrule
\end{tabular}
\end{table}

\paragraph{Aggressive-dose sweep.} The route-mediated share rises from $31\%$ (matched
$1\times$) to $58\%$ near $5.4\times$ dose, then significantly declines to $54\%$ at
$11.7\times$ as the direct KV-corruption channel overtakes routing (channel crossover:
direct $\times2.35$ vs route $\times2.03$; OLMoE, $N{=}384$ paired). The
aggressive-regime recovery ceiling for any routing-consistency intervention is bounded by
the \emph{peak} share $\approx0.58$. This peak is a simulated-injector measurement; the
real-kernel ladder (\S\ref{sec:extval}) shows the real share \emph{declines} at aggressive
bits, so the fake-quant curve likely over-states it there. The strict-monotone prediction
is a registered miss (the $5.4\times{\to}11.7\times$ decline is significant, paired
$+0.0339$ $[+0.0210,+0.0464]$ at $N{=}384$).

\paragraph{Symmetric attenuation and directional recovery.} Direction-agnostic
boundary-smoothing nets $\approx 0$ with a small significant residual ($+0.00247$
$[+0.00046, +0.00458]$ nats): on harmful-flip tokens $\Delta=+0.043$ $[+0.037, +0.049]$, on
beneficial-flip tokens $\Delta=-0.047$ $[-0.055, -0.039]$---near-equal, oppositely signed,
so symmetric attenuation cancels under sign-balanced flip harm. This is the mechanism
behind the class taxonomy: a direction-agnostic (``class-B'') remedy is at the floor, while
a directional clean-reference (``class-C'') remedy recovers the bounded slice above.
\emph{Instrument firewall:} the boundary-smoothing equivalence result (a fixed-process
exploratory $5.1\%$ equivalence region on the harmful component) is a \emph{different
instrument} from the clean-route-pin recovery ($8.9\%$ on OLMoE / $8.4\%$ on DeepSeek) and
from the $0.456$ clean-reference remedy on DeepSeek---three different measurements, never
conflated.

\paragraph{Exchange-rate arm.} Storing clean routes ($\sim 22$ B/token/layer) is
$\approx 28.7\times$ more byte-efficient (recovery-weighted) than a $4{\to}8$-bit prefix-KV
upgrade ($\sim 2048$ B/token/layer) at recovering route-mediated damage---but \emph{partial,
not sufficient}: routes recover $0.291$ of the route ceiling against the precision arm's
$0.946$ (an upper bound on its route-mediated recovery), so the pre-registered ``routes
$\ge 0.5\times$ precision'' bar is not met. The robust statement is the per-byte
efficiency, not a dominance claim.

\begin{table}[t]\centering\small
\caption{Predictions registry (pre-registered; a way to lose, a mechanism citation, and a
named settling measurement per entry).}
\label{tab:predictions}
\begin{tabular}{@{}c p{4.6cm} p{4.5cm}@{}}
\toprule
P & Prediction & Status / planned evaluation \\
\midrule
P1 & Jump-eliminating mechanism removes $\approx0.998\times$RMF; residual $=$ jitter & planned (twin-gate test) \\
P2 & Static W/A logit alignment repairs W/A, \emph{not} KV & planned; precision ladder consistent \\
P3 & Trained vanishing-boundary gates: RMF $\approx$ jitter share & planned \\
P5 & Window/CUSUM residual cannot beat the benefit floor & reserved \\
P7 & Normalization/reference semantics moderate prefix-reference recovery & association; \textbf{controlled flag-swap done} (\S\ref{sec:crossmodel}): re-scoped to a damage-magnitude moderator, \emph{not} a recoverability mechanism \\
P8 & Channel-crossover generality (dose-graded share) & planned (dose-graded sweep) \\
\bottomrule
\end{tabular}
\end{table}

\begin{table}[t]\centering\small
\caption{Negative-result ledger (symmetric reporting; internal conjectures held to the
same standard as external claims).}
\label{tab:ledger}
\begin{tabular}{@{}p{4.5cm} p{4.4cm} p{4.4cm}@{}}
\toprule
Discarded & Evidence & Reason for rejection \\
\midrule
Strict held-out impossibility exclusion & oracle recovery $+0.0042$ $[0.0003,0.0082]$ on the reserved test split & upper CI $0.0082$ exceeds the pre-registered $0.008$ bar --- exclusion \emph{unconfirmed} out of sample (the tiny-recovery direction holds) \\
Sequence-granularity selective control & split-half tie & paired CI included zero \\
Trained-predictor detection & budget-matched replay & paired-vs-random CI included zero \\
``Amplifying cascade'' headline & relative-perturbation re-read & scale artifact $\to$ event accumulation \\
``Flux residual'' channel & attribution re-run & was nonlocal jumps (flux $\approx0.2\%$) \\
Adaptive-$k$ attenuation & re-run $-0.006$ nats & at the $\sigma_{\mathrm{proc}}$ floor \\
Naive route-hold & re-run $-0.95$ nats & stale-reference catastrophe, not class-B evidence \\
\bottomrule
\end{tabular}
\end{table}

\begin{table}[t]\centering\small
\caption{Evidence matrix (the primary splits are calibration; the reserved test split was read once, pre-registered, \S\ref{sec:extval}. FD =
full-decode pin, pfx = prefix-only reference).}
\label{tab:evidence}
\resizebox{\linewidth}{!}{%
\begin{tabular}{@{}llccccc@{}}\toprule
Model & Disturbance & $N$ & RMF & Harm-sign & Clean pin & Smoothing \\ \midrule
OLMoE-1B-7B & sim.\ KV4 & 96/384 & $0.31$ & yes & yes & yes \\
Qwen1.5-MoE-A2.7B & sim.\ KV4 & 96 & $0.29$ & --- & weak & --- \\
DeepSeek-MoE-16B & dose-matched KV4 & 384 & partial & yes & yes & --- \\
Qwen3-30B-A3B & sim.\ KV4 & 384 & --- & --- & pfx null; FD $+0.408$ & --- \\
OLMoE (weight domain) & expert W3/W4 & --- & LB $0.115$--$0.123$ & --- & yes & --- \\
\bottomrule
\end{tabular}}
\end{table}

\end{document}